\def\year{2020}\relax
\documentclass[letterpaper]{article}
\usepackage{aaai2020}
\usepackage{times}
\usepackage{helvet}
\usepackage{amsmath}
\usepackage{courier}
\usepackage{url}
\usepackage{graphicx}
\usepackage{mathrsfs}
\usepackage{amsfonts}
\title{Learning to Compare for Better Training and Evaluation of Open Domain Natural Language Generation Models}
\author{Wangchunshu Zhou, ~~~ Ke Xu\\
State Key Lab of Software Development Environment, Beihang University, Beijing, China\\
{zhouwangchunshu@buaa.edu.cn, kexu@nlsde.buaa.edu.cn}
}
\begin{document}
\maketitle
\begin{abstract}
\begin{quote}
Automated evaluation of open domain natural language generation (NLG) models remains a challenge and widely used metrics such as BLEU and Perplexity can be misleading in some cases. In our paper, we propose to evaluate natural language generation models by learning to compare a pair of generated sentences by fine-tuning BERT, which has been shown to have good natural language understanding ability. We also propose to evaluate the model-level quality of NLG models with sample-level comparison results with skill rating system. While able to be trained in a fully self-supervised fashion, our model can be further fine-tuned with a little amount of human preference annotation to better imitate human judgment. In addition to evaluating trained models, we propose to apply our model as a performance indicator during training for better hyperparameter tuning and early-stopping. We evaluate our approach on both story generation and chit-chat dialogue response generation. Experimental results show that our model correlates better with human preference compared with previous automated evaluation approaches. Training with the proposed metric yields better performance in human evaluation, which further demonstrates the effectiveness of the proposed model.
\end{quote}
\end{abstract}

\section{Introduction}

Recent advances in sequence-to-sequence learning architecture~\cite{sutskever2014sequence} and the transformer model~\cite{vaswani2017attention} have raised increasing interest in natural language generation (NLG) tasks, including story generation~\cite{fan2018hierarchical}, open-domain dialogue response generation~\cite{Sordoni_2015} and abstractive summarization~\cite{See_2017}. Despite the fast advances of models, there remains a huge gap in the evaluation of NLG models and it is hard to measure the progress due to the lack of good evaluation metrics. While perplexity is a good measure of how well a model fits some data, it does not measure performance at the desired task. Word overlap based metrics such as BLEU~\cite{papineni2002bleu}, METEOR~\cite{banerjee2005meteor} and ROUGE~\cite{lin2004rouge} capture quality better than the perplexity and are useful in translation and summarization. However, they still correlate poorly with human evaluation~\cite{liu2016not} in open domain text generation tasks including story generation and dialogue response generation because two equally good generated texts may have no n-gram overlap. Human evaluation is generally considered to be the gold standard evaluation, however, it does not scale well as it is generally expensive and time-consuming to conduct human evaluation.

Apart from measuring relative progress between different models, automated evaluation metrics also play an important role in the training stage of NLG models. It is a common practice to tune the model hyperparameter, detect convergence, perform early-stopping, and select the best checkpoints based on the model's performance on automated evaluation metrics. While acceptable for tasks where automated metrics correlate well with human evaluations, including machine translation and text summarization, this can be erroneous and result in sub-optimal training in open domain NLG tasks because available automated metrics correlate poorly with human evaluation, as demonstrated in the experimental section of this paper.

To tackle the aforementioned problems, in this paper, we propose a self-supervised approach with transfer learning to learn to compare the quality of two samples as an automated comparative Turing test. The motivation of our approach is that we can better assess the quality of generated samples or trained NLG model by comparing it with another one. Our model is a text pair classification model trained to compare the task-specific quality of two samples, which is then used to evaluate the quality of trained NLG models. As human preference annotation is generally expensive, our model is designed to be able to perform self-supervised training using only generated samples and gold reference samples without human preference annotation. When human preference annotation is available, our model can be further fine-tuned to better imitate human judgment. To evaluate the model-level quality of NLG models based on pairwise comparison in sample-level, we adopt the skill rating system similar to ELO~\cite{elo1978rating} and Trueskill~\cite{herbrich2007trueskill}, which is a method for assigning a numerical skill to
players in a player-vs-player game, given a win-loss record of games played. In our scenario, the players are NLG models to be evaluated and a higher rating indicates a better model. The skill rating system makes it possible to evaluate all n models without needing to run $n^{2}$ matches and is able to take into account the amount of new information each comparison provides.   

The contribution of this paper is threefold:
\begin{itemize}
    \item We propose a ``learning to compare'' model to better assess the quality of text generated by NLG models based on pairwise comparison. Our model is able to transfer natural language understanding knowledge from BERT by fine-tuning in a self-supervised way while also able to be further fine-tuned with human preference annotation. Once trained, our model is able to perform inter-model comparison without the need for gold references, which greatly enlarges the potentially available test set and reduces the potential risk of overfitting the reference in the test set.
    \item We propose to use the skill rating system to perform model-level evaluation based on the sample-level evaluation information provided by our pairwise comparison model. The skill rating system is more efficient and accurate than several baseline approaches.
    \item We conduct experiments on both story generation task and open domain dialogue response generation task. Experimental results show that our approach correlates better with human evaluation on both datasets. Moreover, we show that using automated metrics such as BLEU to perform hyperparameter tuning and early-stopping results in sub-optimal model and our approach helps alleviate this problem.
\end{itemize}

\section{Related Work}

Evaluation of NLG models has been a long-standing open problem. While human evaluation may be ideal, it is generally expensive to conduct and does not scale well. Various automated evaluation approaches are proposed to facilitate the development and evaluation of NLG models. We summarize these evaluation approaches below.

\textbf{Text Overlap Metrics}, including BLEU~\cite{papineni2002bleu}, METEOR~\cite{banerjee2005meteor} and ROUGE~\cite{lin2004rouge}, are the most popular metrics employed in the evaluation of NLG models. They evaluate generated text by comparing the similarity between the generated text and human written references. While this works well in tasks where the diversity of acceptable output is limited, such as machine translation and text summarization, text overlap metrics are shown to have weak or no correlation with human judgments in open domain natural language generation tasks~\cite{liu2016not}. There are two major drawbacks in these metrics. First, text overlap metrics can not distinguish minor variations in a generated text which may make the sentence not equally grammatically correct or semantically meaningful. Second, there may exist multiple equally good outputs for the given input and comparing against one gold reference can be erroneous. 

\textbf{Perplexity} is commonly used to evaluate the quality of a language model. It measures how well a probability distribution predicts a sample and captures the degree of uncertainty in the model. It is used to evaluate models in open-domain NLG tasks such as story generation~\cite{fan2018hierarchical} and open domain dialogue systems. However, ``how likely a sentence is generated by a given model'' may not be comparable across different models and does not indicate the quality of the sentence.    

\textbf{Parameterized Metrics} learn a parameterized model to evaluate generated text. Adversarial evaluation models~\cite{kannan2017adversarial,li2017adversarial} assigns a score based on how easy
it is to distinguish the dialogue model responses from human responses. However, training such a discriminator can be difficult as the binary classification task can be easily over-fitted and leads to poor generalizability~\cite{kannan2017adversarial}. Moreover, the information we get from the discriminator accuracy is limited as we can not compare the quality of two generated sentences when they both succeed or fail in fooling the discriminator. Recent study shows that the discriminator accuracy does not correlate well with human preference~\cite{garbacea2019judge}. Automated Dialogue Evaluation
Model (ADEM)~\cite{lowe2017towards} is another parameterized metric proposed for dialogue system evaluation. It learns to score a generated dialogue response based on the context and the human written reference. However, it requires human-annotated scores for generated sentences. It is generally hard to design appropriate questions for crowdsourcing these scores, which makes the annotation very expensive to get and the inter-annotator agreement score is only moderate~\cite{lowe2017towards}. As a result, the training data is limited and noisy, which makes the scoring task even harder. It can be problematic when comparing models with similar quality. In addition, this model is designed only for evaluating dialogue response generation models. More recently, embedding similarity based metrics such as HUSE~\cite{shimanaka2018ruse} and BERTScore~\cite{zhang2019bertscore}. These metrics alleviate the first problem of text overlap metrics by modeling semantic similarity better. However, they can not address the response diversity problem and thus are only suitable for machine translation and text summarization.

Another line of research on NLG evaluation is to unify human evaluation with statistical evaluation~\cite{hashimoto2019unifying,chaganty2018price}. These works are orthogonal to our paper as they mainly focus on the combination of human evaluation and automated evaluation.

Another related work of our research is the skill rating system, which evaluates players by observing a record of wins and losses of multiple players and inferring the value of a latent, unobserved skill variable for each player that explains the records of wins and losses. It is first adopted to evaluate GANs~\cite{goodfellow2014generative} for synthesizing images~\cite{olsson2018skill} by competing generators against discriminators. Their approach is an approximation of skill rating as the original skill rating system requires game played by two symmetric players, while in their system the players are asymmetric. Their approach does not include the ``tie'' option, thus can not distinguish cases where the discriminator is confident enough or not. More importantly, their approach is only designed for evaluating GANs while our approach can be used for any NLG models.

\section{Methodology}

We present the proposed approach in this section. We begin with the sample-level pairwise comparison model. Afterwards, we introduce how to adopt the skill rating system to perform model-level evaluation of NLG models.

\subsection{Learning to Compare}

\begin{figure*}
    \centering
    \includegraphics[width=1.25\columnwidth]{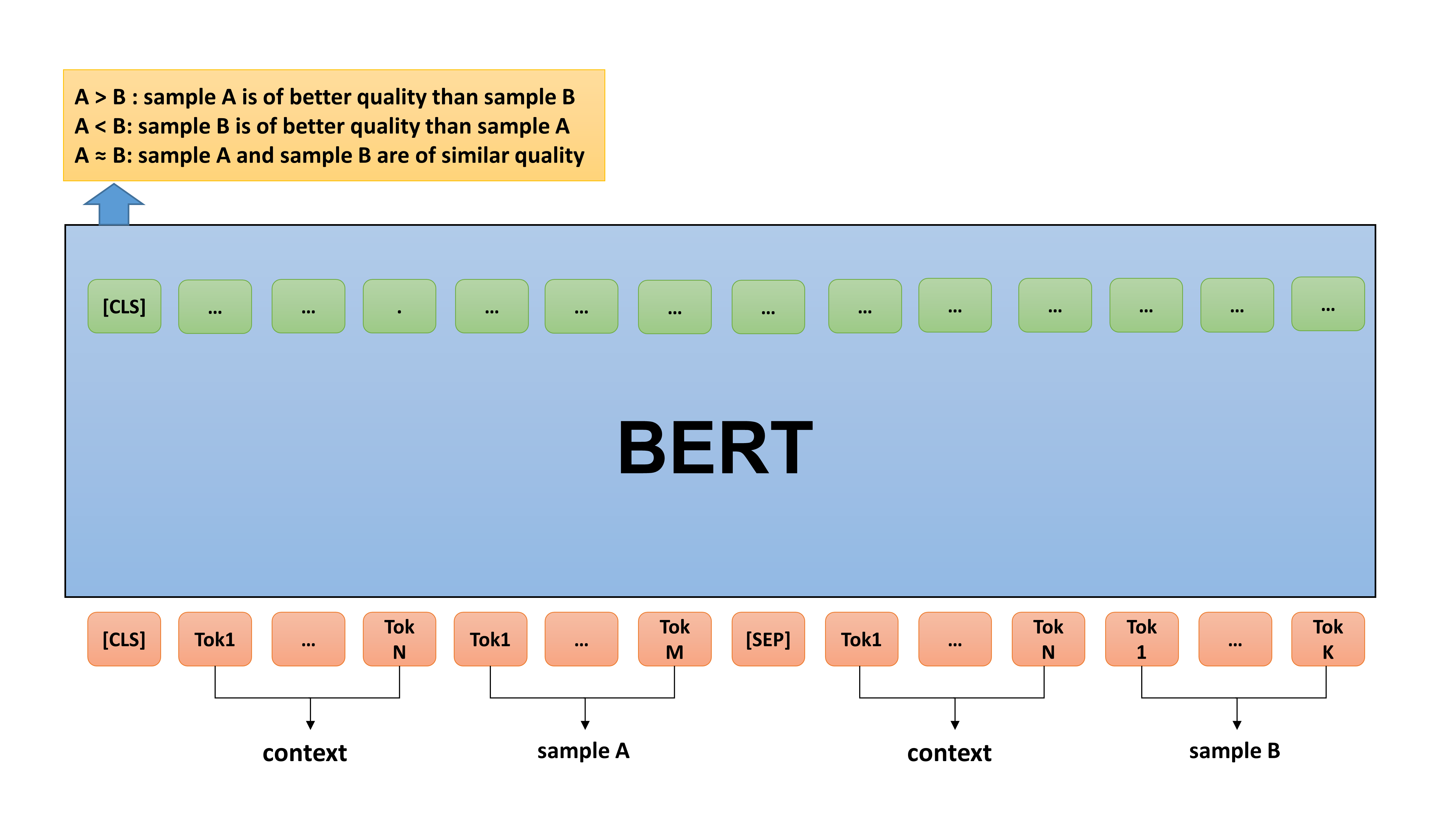}
    \caption{model architecture of the comparative evaluator, the context is concatenated with generated samples.}
    \label{fig:my_label}
\end{figure*}

The proposed comparative evaluator is a text pair relation classifier which is trained to compare the task-specific quality of two samples. The motivation of evaluating one sample by comparing it with another sample is drawn from the insight learned when conducting human evaluation for NLG models. We find that when comparing two NLG models, instead of asking human annotator to assign scores separately for samples generated by different models, which resembles the case in the ADEM model~\cite{lowe2017towards}, it is much easier for human annotators to directly compare one sample generated by the first model against another sample from the second model pairwisely and compute the win/loss rate. The comparison-based evaluation may also be more accurate, which is demonstrated by a higher inter-annotator agreement score in our preliminary experiments.  

The comparative evaluator learns a total order of sample quality by classifying whether the first compared sample is better ($>$), worse ($<$), or indistinguishable ($\approx$) in terms of its quality compared with another sample. In this way, our model encodes the inductive bias that sometimes two samples can have similar quality and it is hard and unreliable to choose the better sample. By giving our model the third ``tie'' option, it can explicitly express its uncertainty and choose its preference only when being confident enough. This design choice is motivated by the practice that adding the ``tie'' option for human annotator when performing pairwise human evaluation can often make the comparison easier and more reliable. For a text sample, our comparative evaluator can provide a more informative assessment than the binary discriminative evaluator because one evaluated sample can receive multiple feedback from the comparative evaluator by comparing it with multiple other samples. In contrast, the discriminative evaluator can only evaluate a sample once, which is more likely to suffer from the inherent uncertainty of the evaluator.

We propose two approaches to construct pairwise training examples for training a comparative evaluator. The first approach generates \textbf{strong supervision examples}. It is based on the intuition that human written references are generally of better quality than machine-generated samples, and it is hard to tell the difference in term of the quality when two compared samples are both generated by machines or human written reference. We denote $S_{+}$$/$$S_{-}$ as the set of real/generated samples. For a real sample $s_{+}\in S_{+}$ and a generated sample $s_{-}\in S_{-}$, we assign the label ``better ($>$)'' to the pair ($s_+$, $s_-$) and ``worse ($<$)'' to ($s_-$, $s_+$). For two samples both from real data or from the generated samples, we assign the label ``indistinguishable ($\approx$)'' to such pairs (i.e., ($s_+^i$, $s_+^j$) and ($s_-^i$, $s_-^j$)). For a training set with $n$ real samples and $n$ generated samples, we can construct $\binom{2n}{2}$ pairwise training examples for the comparative evaluator, allowing to enhance the generalization ability and introduce more informative learning signals than the standard real/fake binary discriminative evaluator. Note that when constructing a sample pair ($s_-^i$, $s_-^j$), $s_-^i$ and $s_-^j$ are sampled from the same checkpoint of the same model in order to ensure that they are of similar quality in expectation. 

One problem of the strong supervision approach is that it always labels two generated samples as indistinguishable. However, during inference, the input of the comparative evaluator is a pair of two generated samples from different models. Thus it requires the model to capture the quality relation in training examples and generalize well to successfully compare two samples rather than simply classifying them as indistinguishable, which provides relatively less information for evaluating NLG models.

To tackle this problem, we propose an approach to construct \textbf{weak supervision examples} for training the comparative evaluator. The intuition of our weak supervision approach is that during training, the quality of the NLG model keeps improving until convergence. Given two checkpoints of the same model, we can thus consider samples generated by the more recent checkpoint are of better quality compared with samples generated by the earlier version of the same model. This approach is considered to be weak supervision because the model quality may not improve monotonically and sometimes it is hard to decide whether the model begins to overfit the training data and its quality starts to decline. To minimize the noise introduced by these problems, we empirically set the minimal margin between two selected checkpoints to be $10\%$ of the total training iteration and do not select two ``almost converged'' checkpoints. The construction of training samples is similar to the first approach. In addition, motivated by the fact that the larger the margin between the quality two selected version of the model, the easier for the comparative evaluator to learn to distinguish the training examples, we propose to use curriculum learning~\cite{bengio2009curriculum} by feeding the comparative evaluator with sample pairs with larger margin (i.e. more training iterations between two selected checkpoints) during initial training stage and gradually decrease the margin to let the model gradually learn to capture smaller quality differences. Moreover, when human preference annotation is available, we can additionally fine-tune the comparative evaluator with human annotations.

The comparative evaluator is trained with maximum likelihood estimation (MLE) objective, as described in eq \ref{Discriminator}
\begin{equation}
\label{Discriminator}
 L = -\mathbb{E}_{(\boldsymbol{x}_{1},\boldsymbol{x}_{2})\sim \mathcal{X}} [{\log {D^{Q(\boldsymbol{x}_1,\boldsymbol{x}_2)}_{\phi}(\boldsymbol{x}_{1},\boldsymbol{x}_{2})}}] 
\end{equation}
where $\mathcal{X}$ is the set of pairwise training examples contructed as described above, $Q(\boldsymbol{x}_1, \boldsymbol{x}_2) \in \{>,<,\approx\}$ is the true label for the pair ($\boldsymbol{x}_1$, $\boldsymbol{x}_2$), $D_\phi^q(\boldsymbol{x}_1, \boldsymbol{x}_2)$ is the probability of the comparative discriminator's prediction being $q$ ($q \in \{>,<,\approx\}$) for the pair ($\boldsymbol{x}_1$, $\boldsymbol{x}_2$).

As comparing the quality of generated text requires good natural language understanding ability and our comparative evaluator is formulated as a sentence pair classification model, we propose to fine-tune  BERT~\cite{devlin2018bert} as the comparative evaluator, the architecture of the resulting comparative evaluator is illustrated by Figure 1. Note that the compared sample A and B are based on the same context, which ensures that they are comparable.

\subsection{Skill Rating}

In player-vs-player games such as chess or tennis, skill rating systems such as Elo~\cite{elo1978rating} or Glicko2~\cite{glickman2012example} evaluate players by observing a record of wins and losses of multiple players and inferring the value of a latent, unobserved skill variable for each player that explains the records of wins and losses. We adopt the skill rating system for model-level evaluation of NLG models. By taking the trained comparative evaluator as the ``playground'' and NLG models as ``player'', the ``player-vs-player'' game is played by sampling one output sample from each NLG model conditioning on the same input and the game output is decided by the comparative evaluator. 

Following previous work~\cite{olsson2018skill}, in our paper, we use the Glicko2 system~\cite{glickman2012example}. The employed system can be summarized as follows: each player's skill rating is represented as a Gaussian distribution, with a mean and standard deviation, representing the current state of the evidence about their ``true'' skill rating. As we evaluate frozen snapshots of NLG models, we disabled an irrelevant feature of Glicko2 that increases uncertainty about a human player’s skill when they have not participated in a match for some time. Another difference is that conventional skill rating systems do not support the ``tie'' option, which is important for the system to be stable and reliable in our case because the evaluator is not perfect. To incorporate this feature, we follow the intuition that a player's skill rating should be increased when it draws with another player with a higher skill rating and vice versa. We come up with a simple rule which increases/decreases the skill rating of one player by a ratio (e.g. 0.1) of the changes in its skill rating when it wins/loses if it draws with another player with higher/lower skill rating. In our experiments, the skill rating is performed by randomly sampling two compared models, simulating a ``game'' between two selected models by sampling one sample from each model and comparing them with the comparative evaluator, and then updating the skill rating of selected models according to the outcome. This procedure is performed iteratively until convergence, which is defined as the order of skill ratings of compared models keeps the same after each model is selected at least 50 times. While the sampling procedure can be optimized by bayesian optimization~\cite{snoek2012practical} or multi-armed bandit algorithms~\cite{vermorel2005multi}, we choose to keep the method as simple as possible and use random sampling. 

\section{Experiments}

We set up experiments in order to answer the following research questions:

\begin{itemize}
    \item \textbf{RQ1}: Can the comparative evaluator correlate better with human preference in sample-level than previous automated metrics when evaluating open domain NLG models? 
    \item \textbf{RQ2}: Can the comparative evaluator correlate better with human preference in model-level, so that our approach can measure the progress on open domain NLG better?
    \item \textbf{RQ3}: As existing approaches fail to correlate well with human preference, whether and to what extent this problem affects the quality of the final NLG model when performing hyperparameter search and early-stopping?
    \item \textbf{RQ4}: If the previous problem exists, can proposed comparative evaluator reduce this problem?
\end{itemize}

\subsection{Experimental Settings}

\subsubsection{Datasets}

We evaluate the effectiveness of the proposed approach on two open domain natural language generation tasks: story generation and open domain dialogue response generation. For story generation, we use the WritingPrompts dataset released by~\citeauthor{fan2018hierarchical}. The WritingPrompts dataset is a large dataset of 303,358 human-generated stories paired with writing prompts from an online forum. NLG models are trained by taking writing prompts as input and generating the whole story. The average length of prompts is 28.4 and the average length of stories is 734.5 words, which makes human evaluation very expensive and better automated metrics are thus critical. For open domain dialogue response generation task, we use the Dailydialog dataset~\cite{li2017dailydialog}, which consists of dialogues that resemble daily conversations across multiple topics. It comprises of 13k dialogues with an average of 7.9 turns per dialog.

\subsubsection{Compared Models and Metrics}

As our objective is to evaluate the evaluators rather than comparing state-of-the-art models, we choose three representative sequence-to-sequence architectures: LSTM~\cite{hochreiter1997long} seq2seq, Convolutional seq2seq~\cite{gehring2017convolutional}, and transformer~\cite{vaswani2017attention} model. We compare models with different architectures, hyperparameter choices, and early-stopping criteria with different automated metrics, as well as human evaluation.

Regarding the evaluation metric (and criteria for choosing hyperparameter choice and early-stopping), we compare the proposed approach with the discriminative evaluator, BLEU score (average of 2-, 3-, 4-grams), perplexity, and ADEM. When evaluating generated stories, we cut off the story at the nearest sentence for stories longer than 250 words.

The proposed comparative evaluator is employed for choosing hyperparameter by performing skill rating among all models trained with different hyperparameter choices\footnote{For each model, we randomly sample 5 hyperparameter choices in a predefined range.}. For early-stopping, as incrementally performing skill rating is computationally expensive, we propose to perform n (e.g. 1000) pairwise comparison between the samples generated by the latest checkpoint and the previous k (e.g. 2) checkpoints and stop training when the wining rate of latest checkpoint keeps being smaller than its losing rate for 5 iterations.

\subsubsection{Detail of Parameterized Evaluators}

The proposed comparative evaluator is trained by fine-tuning BERT-large as a sentence-pair classifier. To ensure fair evaluation, we also train the discriminative evaluator by fine-tuning BERT. For ADEM, we adopt its original implementation as its architecture is relatively complicated. In addition, we perform ablation study by evaluating three variants of the comparative evaluator where it is trained without strong supervision examples, without weak supervision examples, without fine-tuning with human preference annotations, and without transferring from BERT.

\subsubsection{Human Evaluation Procedure}

As human evaluation is expensive, sample-level evaluation is performed jointly with model-level evaluation, which is also used for evaluating the ability of different metrics for performing hyperparameter search and early-stopping. Concretely, we perform 10 groups of evaluations for performing hyperparameter selecting and early-stopping with five compared automated metrics. In each evaluation, each of the five compared metrics is used to select the best hyperparameter combination or early-stopping checkpoint with other variants fixed. 
 
We choose to perform score-based human evaluation for four reasons: 1) the ADEM baseline requires human-annotated score as training examples, 2) we can construct up to $\binom{2n}{2}$ training examples for our comparative evaluator with $n$ human-annotated scores, 3) score-based human evaluation facilitates the evaluation of correlation scores, and 4) as all other metrics do not perform pairwise comparison, using pairwise human evaluation will likely be biased toward our approach. 

We sample 20 generated samples from each model (out of 5) of the 20 evaluation groups. We invite 20 human annotators which are all graduate students with good English language proficiency to score these samples. Each annotator scores one sample from each model, such that each model is uniformly evaluated. The score scales from 1 to 5, higher score indicates better overall sample quality. According to experimental results from~\citeauthor{lowe2017towards}, we do not ask annotators to provide specific scores for fluency or informativeness. To test the inner-annotator agreement score, we additionally ask them to evaluate another 40 generated samples, of which 20 samples are scored from 1 to 5 and another 20 are evaluated based on pairwise comparison with 4 other generated samples and scored to 1-5 based on how many times they are considered to be better than a reference sample. We get an inter-annotator agreement score $\kappa=0.53$ for directly scoring and $\kappa=0.76$ with pairwise comparison, which validates our intuition that evaluation by comparison may be more accurate. These additional human annotations are used as training data for ADEM and the comparative evaluator.

\begin{table*}[t!]
\begin{center}
\resizebox{1.85\columnwidth}{!}{
\begin{tabular}{lcccc}
\hline\hline
 \textbf{Task} & \multicolumn{2}{c}{\textbf{Story Generation}} & \multicolumn{2}{c}{\textbf{Dialogue}}\\ \hline
\textbf{Metric} & \textbf{Spearman} & \textbf{Pearson} & \textbf{Spearman} & \textbf{Pearson} \\ \hline
\bf our approach &  & & & \\
~ - with skill rating  & \bf 0.392($<0.001$) & \bf 0.387($<0.001$) &  \bf 0.473($<0.001$) & 0.469($<0.001$) \\
~ - with random sampled reference & 0.389($<0.001$) & 0.378($<0.001$) & 0.461($<0.001$) & \bf 0.472($<0.001$) \\
ADEM & 0.162 & 0.148 & 0.353 & 0.341 \\
Adversarial Evaluator & 0.105 & 0.111 & 0.197 & 0.182  \\
BLEU & 0.032 & 0.028 & 0.053 & 0.061 \\
Perplexity & 0.052 & 0.057 & 0.024 & 0.015  \\
\hline\hline
\end{tabular}}
\end{center}
\caption{ Sample-level correlation between metrics and human judgments, with p-values shown in brackets. }
\label{resultsample}
\end{table*}

\begin{table*}[t!]
\begin{center}
\resizebox{1.85\columnwidth}{!}{
\begin{tabular}{lcccc}
\hline\hline
 \textbf{Task} & \multicolumn{2}{c}{\textbf{Story Generation}} & \multicolumn{2}{c}{\textbf{Dialogue}}\\ \hline
\textbf{Metric} & \textbf{Spearman} & \textbf{Pearson} & \textbf{Spearman} & \textbf{Pearson} \\ \hline
\bf our approach &  & & & \\
~ - with skill rating  & \bf 0.612($<0.001$) & \bf 0.631($<0.001$) &  \bf 0.764($<0.001$) & \bf 0.783($<0.001$) \\
~ - averaged sample-level skill rating  &  0.518 &  0.541 & 0.651 & 0.675 \\
~ - averaged reference-based score & 0.473 & 0.482 & 0.634 & 0.653 \\
ADEM & 0.291 & 0.302 & 0.541 & 0.553 \\
Adversarial Evaluator & 0.248 & 0.272 & 0.491 & 0.502  \\
BLEU & 0.096 & 0.103 & 0.217 & 0.293 \\
Perplexity & 0.113 & 0.127 & 0.071 & 0.083  \\
\hline\hline
\end{tabular}}
\end{center}
\caption{Model-level correlation between metrics and human judgments, with p-values shown in brackets.}
\label{resultmodel}
\end{table*}

\subsection{Experimental Designs \& Results}

\subsubsection{RQ1: Sample-Level Correlation} To test the correlation of different automated metrics with respect to human preference, we employ different metrics to score the collected 2000 samples and calculate their Pearson and Spearman correlation with human scores. For comparative evaluator, as the evaluation is performed pairwisely and no absolute score is available, we use two different approaches to get an absolute score for each sample: 1) we sample 50 common references from machine-generated samples for each task and compare each sample with all references by the comparative evaluator. A sample gets 3 points when beats a reference, 1 point when draws with the reference, and get 0 point when loses, 2) we adopt skill rating system by regarding each sample as an NLG model which always outputs the same sample and use the skill rating for each sample as its score. To ensure the computational budget to be roughly the same, we fix the number of plays in skill rating to 10,000. 

The experimental results are summarized in Table 1. We can see that the proposed comparative evaluator correlates far better with human judgment than BLEU and perplexity. When compared with recently proposed parameterized metrics including adversarial evaluator and ADEM, our model consistently outperforms them by a large margin, which demonstrates that our comparison-based evaluation metric is able to evaluate sample quality more accurately. In addition, we find that evaluating generated samples by comparing it with a set of randomly selected samples or using sample-level skill rating performs almost equally well. This is not surprising as the employed skill rating is able to handle the inherent variance of players (i.e. NLG models). As this variance does not exist when we regard a sample as a model which always generates the same sample.

\subsubsection{RQ2: Model-Level Correlation} As for model-level evaluation, we employ the average score of the evaluated 100 samples as each model's score and calculate their correlation with human scores. For comparative evaluator, we propose three different approaches to get an absolute score for each sample: 1) we calculate the average reference-based score (method 1 for sample-level comparison) of each sample as model-level score, 2) we calculate the average skill rating of each sample obtained in the experiments of RQ1 as model-level score,  2) we use the proposed skill rating system to get a model-level skill rating for each compared model.  

Results are shown in Table 2. We can see that the proposed comparative evaluator with skill rating significantly outperforms all compared baselines, including comparative evaluator with averaged sample-level scores. This demonstrates the effectiveness of the skill rating system for performing model-level comparison with pairwise sample-level evaluation. In addition, the poor correlation between conventional evaluation metrics including BLEU and perplexity demonstrates the necessity of better automated evaluation metrics in open domain NLG evaluation.

\subsubsection{RQ3\&4: Automated Metrics for Model Training}

We further investigate the impact of imperfect metrics on training NLG models. As described in the human evaluation procedure, we perform 10 runs to test the reliability of each metric when used to perform hyperparameter tuning and early-stopping respectively. In each run, we select the best hyperparameter combination or early-stopping checkpoint based on each of the five compared metrics. Human evaluation is then employed to identify the best choice. We evaluate the performance of each metric by how many times (out of 10) they succeeded in selecting the best hyperparameter combination or early-stopping checkpoint (out of 4) and the average human-annotated score for their selected models.

\begin{table*}[t!]
\begin{center}
\resizebox{1.45\columnwidth}{!}{
\begin{tabular}{lcccc}
\hline\hline
 \textbf{Task} & \multicolumn{2}{c}{\textbf{Hyperparameter Search}} & \multicolumn{2}{c}{\textbf{Early-stopping}}\\ \hline
\textbf{Metric} & \textbf{win times} & \textbf{averaged score} & \textbf{win times} & \textbf{averaged score} \\ \hline
\bf our approach  & \bf 8 & \bf 3.14 &  \bf 9 & \bf 3.41 \\ 
ADEM & 5 & 3.03 & 6 & 3.29 \\
Adversarial Evaluator & 4 & 2.92 & 4 & 3.23  \\
BLEU & 2 & 2.75 & 3 & 3.15 \\
Perplexity & 3 & 2.77 & 3 & 3.11  \\
\hline\hline
\end{tabular}}
\end{center}
\caption{ Performance of different metrics in hyperparameter tuning and earlystop checkpoint selecting.}
\label{resultselect}
\end{table*}

The results are shown in Table 3. We can see that conventional automated metrics perform poorly and result in sub-optimal result when performing hyperparameter search and selecting the best performing checkpoints. Converting evaluation metric from BLEU or perplexity to the proposed comparative evaluator can yield non-neglectable improvements without changing model architecture or training objective. While previous work on NLG evaluation mostly focuses on the evaluation stage and does not explore the influence of imperfect metrics during model training, our experiments demonstrate the existence of this problem and that the proposed method can, to some extent, alleviate this problem.    

\subsection{Qualitative Analysis}

\begin{table*}[t!]
\begin{center}
\resizebox{2.1\columnwidth}{!}{
\begin{tabular}{lccc}
\hline\hline
 \textbf{Context} & \textbf{Sample A} & \textbf{Sample B} & \textbf{Output} \\ \hline
Say,Jim,how about going for a few beers after dinner? & I do not know about it. & No, it is not good. & A $<$ B \\ \hline
 I suggest a walk over to the gym where we can meet some friends. & That's a good idea, ok. & No, I do not like to.  &  Tie \\ \hline
What shall we do ? I don't feel like sitting at home. & We can go for a walk. & I suggest staying at home. & A $>$ B \\
\hline\hline
\end{tabular}}
\end{center}
\caption{Examples of comparison results between two generated samples given context. }
\label{example}
\end{table*}

We present several comparison examples in the Dailydialog dataset for qualitative analysis of the proposed comparative evaluator. From the first example, we can see that the comparative evaluator is capable of identifying that generic and dull responses (i.e. ``I don't know'') should be considered as of worse quality. The second example suggests that our approach handles the diversity in possible responses well, as it regards both positive response and negative response as valid responses. Hopefully, these examples may provide us with some insights about why the proposed metric correlates better with human preference.

\begin{table}[t!]
\begin{center}
\begin{tabular}{lcc}
\hline\hline
\textbf{Metric} & \textbf{Spearman} & \textbf{Pearson} \\ \hline
full model & \bf 0.764 & \bf 0.783 \\
w/o comparison  & 0.491 & 0.502 \\ 
w/o tie option &   0.557  & 0.561 \\ 
w/o strong supervision  & 0.697 & 0.703 \\ 
w/o weak supervision  & 0.728 & 0.737 \\
w/o human annotation  & 0.602 & 0.609 \\
w/o BERT & 0.644 & 0.662 \\
\hline\hline
\end{tabular}
\end{center}
\caption{\label{ablated} Model-level correlation between ablated variants and human judgments in the Dailydialog dataset}
\end{table}

\subsection{Ablation Study}

To better understand the proposed comparative evaluator and analyze the relative importance of its different components, we conduct an ablation study with several variants of the proposed model:

\begin{itemize}
    \item \textit{w/o comparison}: Evaluating generated samples without comparison, which degrades to the adversarial evaluation method.
    \item \textit{w/o strong supervision}: Training the comparative evaluator without ``strong supervision'', which models the inductive bias that human written reference samples are generally of better quality compared with that generated by NLG models.
    \item \textit{w/o weak supervision}: Training without ``weak supervision'', which models the inductive bias that the quality of NLG models generally improves during training.
        \item \textit{w/o human preference annotation} Training without human annotated preference data (i.e. only with strong and weak supervision).
    \item \textit{w/o tie option} The variant of comparative evaluator where the model must select the better sample rather than able to admit its uncertainty.
        \item \textit{w/o BERT} The variant where the model is trained from scratch instead of fine-tuning BERT.
\end{itemize}

We evaluate these model variants on the Dailydialog dataset. Results are presented in Table 5. We can see that comparison-based evaluation is very effective as our model correlates much better than adversarial evaluator. The tie option is also very important as it can prevent the comparative evaluator from making uncertain decision and model the inductive bias that samples generated by the same model are generally of similar quality, which may help our model generalize better. As for different sources of training examples, we find that human preference annotation is the most important, which is not surprising. In addition, we find that the proposed weak supervision also helps, but is of smaller relative importance compared with strong supervision. This may be due to the fact that examples constructed by the weak supervision approach may contain a lot of noise. We can also see that our model correlates well with human preference without training with human preference annotation, this is very important in practice as human annotations are not always available. Finally, we find that transferring the natural language understanding ability from BERT to be very important for the final performance.

\section{Discussion and Conclusion}

In this paper, we present a novel comparison-based parameterized automated evaluation metric for evaluating open domain NLG models. The proposed model is based on the intuition that we can better evaluate the quality of a sample by comparing it with other samples. Our model allows the model to admit its uncertainty with the ``tie'' option. We adopt the skill rating system to perform model-level evaluation based on sample-level pairwise comparison.

By transferring pretrained natural language understanding knowledge from BERT and fine-tuning with strong and weak supervision examples and human preference annotations, our model correlates better with human judgment than other compared metrics. In addition, we find that when used as evaluation metrics, conventional metrics such as BLEU and perplexity may affect the training stage of NLG models as they may lead to sub-optimal hyperparameter choice and checkpoint selection. Our model, in contrast, is much more reliable when performing these choices.

\bibliographystyle{named}
\bibliography{AAAI-ZhouW.1159}
\end{document}